\documentclass{article}
\usepackage{spconf,amsmath,epsfig}
\usepackage{amsfonts}
\usepackage{graphicx}
\usepackage{amsmath}
\usepackage{amssymb}
\usepackage{amsfonts}
\usepackage{booktabs}
\usepackage{tikz}
\usepackage{tikz-3dplot}
\usepackage{mathrsfs}
\usepackage{array}
\usepackage{hyperref}

\title{Fus-MAE: A cross-attention-based data fusion approach for Masked Autoencoders in remote sensing}
\name{Hugo Chan-To-Hing, Bharadwaj Veeravalli}
\address{National University of Singapore, Department of Electrical and Computer Engineering}
\begin{document}
%
\maketitle
\begin{abstract}
    \textit{Selected for IGARSS 2024 Oral. Please cite the version on IEEE Xplore.} \\
    Self-supervised frameworks for representation learning have recently stirred up interest among the remote sensing community, given their potential to mitigate the high labeling costs associated with curating large satellite image datasets. In the realm of multimodal data fusion, while contrastive learning methods can help bridge the domain gap between different sensor types, they rely on data augmentation techniques that require expertise and careful design, especially for multispectral remote sensing data. A possible but rather scarcely studied way to circumvent these limitations is to use a masked image modelling based pretraining strategy. In this paper, we introduce Fus-MAE, a self-supervised learning framework based on masked autoencoders that uses cross-attention to perform early and feature-level data fusion between synthetic aperture radar and multispectral optical data - two modalities with a significant domain gap. Our empirical findings demonstrate that Fus-MAE can effectively compete with contrastive learning strategies tailored for SAR-optical data fusion and outperforms other masked-autoencoders frameworks trained on a larger corpus. For replicability, code and weights are provided in this \href{https://github.com/hugocth/fus-mae}{github} repository.
\end{abstract}
\section{Introduction}
\label{sec:intro}

Multi-modal learning has been attracting increasing attention over the past years, for a vast array of modalities such as RGB-Depth \cite{song2015sun} or text-image \cite{radford2021learning}. In particular, recent research has
established theoretical justifications for a performance edge of deep multi-modal learning over unimodal \cite{huang2021makes}. Within the domain of data fusion for remote sensing (RS), two modalities are extensively studied: synthetic aperture radar (SAR) and optical imagery. Indeed, these modalities inherently complement each other: while SAR data offers all-weather and cloud-penetrating capabilities, it suffers from speckle noise, rendering its interpretation challenging. On the other hand, optical data, though subject to weather and seasonal constraints, proposes natural-looking (e.g. RGB) and less noisy images, facilitating interpretation. Hence, their combination proves relevant for tasks such as land cover classification, and opens doors to applications like cloud removal \cite{ebel2022sen12ms} and SAR despeckling \cite{vitale2019guided}. \par

Self-supervised learning (SSL) has stirred substantial interest in various machine learning fields, such as natural language processing (NLP) \cite{radford2018improving, devlin2018bert} and computer vision \cite{chen2020simple, he2022masked}. One of its key characteristics is its ability to learn powerful representations without the need for labeled data, which is particularly interesting in the domain of RS, where data annotation can be costly and often requires specific expertise. \par

The increasing availability of large-scale public SAR-optical datasets such as, BigEarthNet-MM \cite{sumbul2021bigearthnet-mm}, SEN12MS \cite{schmitt2019sen12ms} and, more recently, SSL4EO-12 \cite{wang2022ssl4eo} fostered research on SSL approaches for SAR-optical fusion. However, the majority of existing research leans towards contrastive learning \cite{survey-ssl-rs}, which, while effective, presents certain limitations. These include a reliance on data augmentations, which need to be carefully designed to adapt to the specificities of remote sensing images (RSI), as well as the necessity for negative samples, which necessitates a large batch size, hence large compute resources.  \par

Recent advances in masked image modelling (MIM) \cite{he2022masked} set a new state-of-the-art for some visual representation learning tasks. Despite MIM avoiding the above-mentioned drawbacks of contrastive learning, to the best of our knowledge, the literature on data fusion for RSI using MIM remains relatively scarce \cite{fuller2022satvit, zhang20233dmae}. In this paper, we explore this alternative pretraining approach for SSL data fusion, with our contributions being summarized as following:
\begin{enumerate}
    \item We introduce Fus-MAE, a self-supervised, MAE-based framework able to perform early-level as well as feature-level data fusion.
    \item We demonstrate empirically that an early-fusion approach leveraging cross-attention is the best pre-training strategy for transformers to perform SAR-optical data fusion tasks.
    \item We show that our Fus-MAE model can compete with some of the most recent contrastive learning approaches tailored for RSI data fusion.
\end{enumerate}

\begin{figure*}[h!]
  \centering
    \includegraphics[scale=0.5]{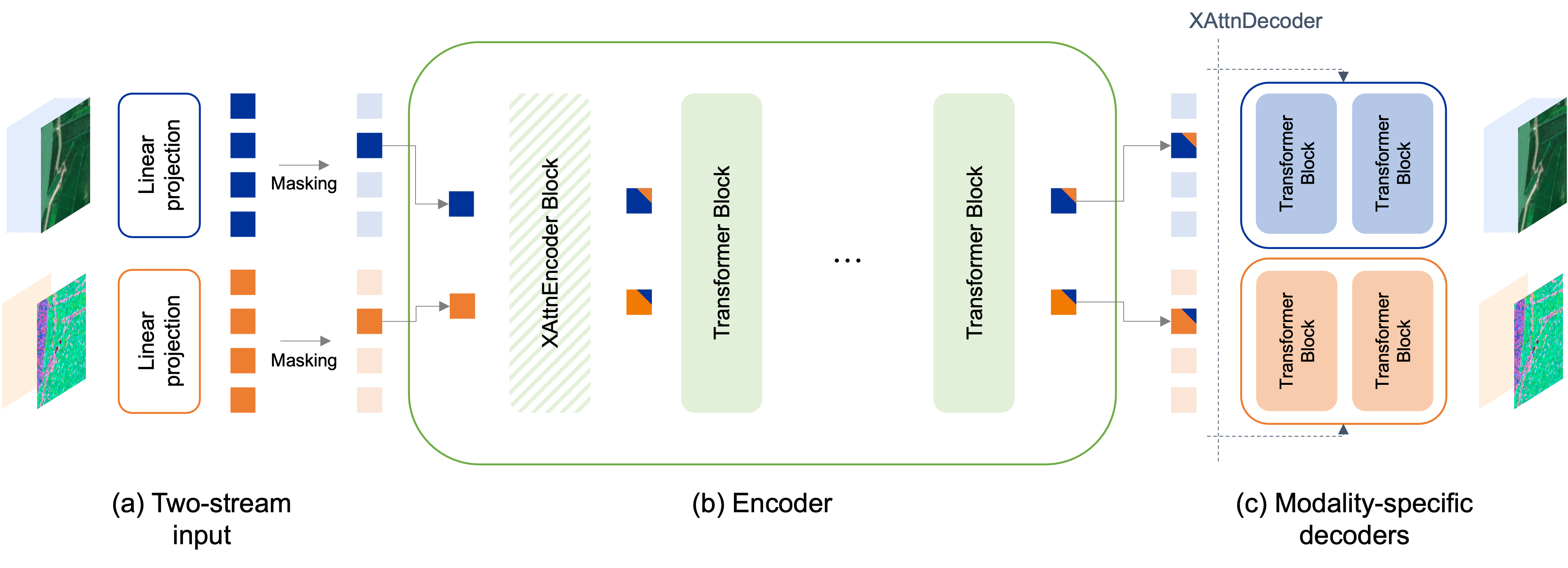}
    \caption{Overall architecture of our Fus-MAE framework.}
    \label{fig:overall architecture}
\end{figure*}

\section{RELATED WORKS}
\label{sec:related works}

\textbf{Self-supervised learning in RS} - As per in the literature \cite{survey-ssl-rs}, self-supervised learning methods can be classified into 3 categories: (1) generative methods, where the pretext task is to reconstruct a corrupted signal at pixel-level (e.g. downsampled \cite{nguyen2021self} or masked \cite{cong2022satmae}), (2) predictive methods, where the objective is to learn semantic context features through pretext tasks such as predicting the relative positions of two patches of an image (for spatial features) \cite{doersch2015unsupervised} or gray-to-RGB coloration (for spectral features) \cite{larsson2017colorization}, and (3) contrastive learning methods, which traditionally aims at creating an embedding space where views of the same instance are drawn closer (positive views), while unrelated views are pulled apart (negative views) \cite{he2020momentum, chen2020simple}. For SAR-optical data fusion, most research efforts lean towards the latter: Chen and Bruzzone \cite{chen2021self} studied early, intermediate and late fusion of SAR and optical images by jointly training two ResUnets with a multiview contrastive loss. Montanaro et al. \cite{montanaro2022semi} used the SimCLR framework to bring the embeddings from different modalities closer. Wang et al. \cite{wang2022self} adapted the knowledge distillation-based DINO framework \cite{caron2021emerging}, which doesn't require negative samples, getting rid of the need for a large batch size. While quite successful, all of these contrastive methods need a careful design of the data augmentation pipeline to create the positive views, whose quality can be difficult to assess. To bypass this challenge, we choose to focus on a generative method which doesn't require data augmentations: masked image modelling. \par

\textbf{Masked image modelling in RS} - He et al. \cite{he2022masked} recently proposed a variation of the denoising autoencoder architecture (DAE), where input images patches are randomly masked with a high masking ratio, leaving only a small subset of patches to be fed into transformer encoder. Then, a shallow decoder reconstructs the image using both obtained latents and masked tokens. Called masked autoencoder (MAE), this framework set a new state-of-the-art on ImageNet-1K, while accelerating training time considerably due to the lower number of processed input tokens and the lightweight decoder. Cong et al. \cite{cong2022satmae} adapted this architecture for optical data by adding multi-domain encoding (e.g. positional+temporal or positional+spectral). Sun et al. \cite{sun2022ringmo} trained an MAE-based model on a 2M optical images dataset and claim to have achieved SOTA performance on various RS datasets. Allen et al. \cite{allen2023large} followed up with a comparable work for SAR images. However, despite recent advancements on masked image modelling for data fusion on the natural domain \cite{bachmann2022multimae}, literature is less extensive for SAR-optical, with some attempts to train MAEs by stacking SAR and optical data along the channel axis \cite{fuller2022satvit} and some studies on specialized masking strategies \cite{zhang20233dmae}. In this paper, we propose some architectural changes to study early, intermediate and late fusion strategies to pave the way for further research.

\section{Methodology}
\label{sec:methodology}

Our work is inspired by MultiMAE, a masked autoencoder-based architecture with a proven track record for natural images, capable of taking different modalities as input \cite{bachmann2022multimae} with its hybrid-stream architecture. In this section, we describe the Fus-MAE architecture by motivating the need for a multi-task encoder in section \ref{subsec:multi-modal encoder} and a multi-task decoder in section \ref{subsec:multi-task decoder}. Two masking strategies are considered, with detailed provided in section \ref{subsec:masking strategies}. The overall architecture is shown in Figure \ref{fig:overall architecture}. \par

\subsection{Multi-modal encoder}
\label{subsec:multi-modal encoder}

As in MAE \cite{he2022masked}, our encoder is a ViT \cite{dosovitskiy2020image} and takes linearly embedded vector representations of patches as input tokens. Let $\textbf{I}_1 \in \mathbb{R}^{H \times W \times C_1}$ and $\textbf{I}_2 \in \mathbb{R}^{H \times W \times C_2}$ be the respective tensor representations of a SAR and an optical satellite image. An intuitive fusion strategy would be to stack SAR and optical RSI data along the channel dimension, and create patches from the obtained tensor. Since this early concatenation technique focuses the entire fusion process onto the single patch projection layer, we hypothesize that it would not be expressive enough to effectively describe cross-modal interactions given the significant domain gap between SAR and optical data. To solve this challenging early fusion task, Fus-MAE replaces the first encoder block by a "Cross-attended patch projection" module, which encodes finer-grained multi-modal information into the input tokens.

\textbf{Cross-attended patch projection} - We first create unimodal tokens using modality-specific patch projection layers. More specifically, given a patch size $P$, for each modality $i$, a 2D convolutional layer $\text{Conv2d}_i$ of kernel size $P \times P$ and stride $P \times P$ is applied, and then positional embeddings $\textbf{E}_{emb}$ are added, to get a set of $(H/P)^2$ tokens $\textbf{z}_{0,i}$:
\begin{equation}
  \textbf{z}_{0,i} = \text{Conv2d}_i(\textbf{I}_i) + \textbf{E}_{emb}
  \label{eq:unimodal_tokens}
\end{equation}
Then, to perform an early fusion operation, we introduce a block called XAttnEncoder (for cross-attention encoder), which is defined as:
\begin{equation}
  \text{fus}(\textbf{x}, \textbf{y}) = \textbf{x} \oplus \textbf{y} + \text{CA}(\textbf{x},\textbf{y}) \oplus \text{CA}(\textbf{y}, \textbf{x})
\end{equation}
\begin{equation}
  \text{XAttnEncoder}(\textbf{x}, \textbf{y}) = \text{fus}(\textbf{x}, \textbf{y}) + \text{MLP}(\text{fus}(\textbf{x}, \textbf{y}))
  \label{eq:xattn-encoder}
\end{equation}
with $\oplus$ the concatenation operation, MLP a two layer feed-forward network with a GELU non-linearity, and CA a cross attention layer defined as:

\begin{align}
  \text{CA}(\textbf{x}, \textbf{y}) &= \text{Attention}(\textbf{Q}_x, \textbf{K}_y, \textbf{V}_y) \\
  &= \text{Softmax}\left(\frac{\textbf{Q}_x {\textbf{K}_y}^T}{\sqrt{d_q}}\right)\textbf{V}_y
  \label{eq:CA}
\end{align}

Our final set of input tokens $\textbf{z}_0$ is obtained by feeding the sets of unimodal tokens to this XAttnEncoder block, as well as appending a global token $z_{CLS}$ with learned embedding, similarly to \cite{bachmann2022multimae}:
\begin{equation}
    \textbf{z}_0 = \text{XAttnEncoder}(\textbf{z}_{0,1}, \textbf{z}_{0,2}) \oplus z_{CLS}
  \label{eq:input_tokens}
\end{equation}
The main idea behind the replacement of the first encoder block with this XAttnEncoder block is that, given the large domain gap between the two modalities, feeding both streams of unimodal tokens into the encoder block's self-attention layer would have resulted in an attention map akin to a block diagonal matrix, with poor cross-modality understanding. On the other hand, the XAttnEncoder block incitates the network to model cross-modal interactions very early, creating tokens with unimodal bias as well as relevant cross-modal information.

\textbf{Modality-biaised latents} - Let $N$ be the depth of our ViT encoder $\mathcal{E}_N$. In the case where the XAttnEncoder block is used, $\mathcal{E}_N$ encoder will be composed of one XAttnEncoder block followed by N-1 Transformer encoder blocks. We feed our unimodal tokens $\textbf{z}_{0,i}$ to the encoder, to obtain a set of modality-biased latents $\textbf{z}_{N}$, which can be decomposed as:

\begin{align}
    \textbf{z}_n &= \mathcal{E}_N(\textbf{z}_{0,1}, \textbf{z}_{0,2})  \\
    &= \textbf{z}_{N,1} \oplus \textbf{z}_{N,2}
\end{align}

\subsection{Multi-task decoder}
\label{subsec:multi-task decoder}

With the aim of performing feature-level data fusion, our architecture proposes to set up one encoder per modality. Following MAE \cite{he2022masked}, we use lightweight decoders, therefore adding decoders does not significantly increase the overall computational complexity of the model. We feed modality-biased latents $\textbf{z}_{N,i}$ to their respective decoder $\mathcal{D}_i$, to obtain a reconstruction of the original RSI data $\hat{\textbf{I}_i}$
\begin{equation}
    \hat{\textbf{I}_i} = \mathcal{D}_i(\textbf{z}_{N,i})
\end{equation}
We then compute the Mean Square Error loss over the reconstructed tokens only, and backpropagate the gradients over the whole architecture. \par

To further insist on feature-level cross-modal information fusion, following MultiMAE \cite{bachmann2022multimae}, we introduce a XAttnDecoder block, which performs cross-attention between the modality-biaised latents, before feeding them to $\mathcal{D}_i$:
\begin{equation}
    {\textbf{z}^{\times}}_{N,i} = \textbf{z}_{N,i} + \text{CA}(\textbf{z}_{N,i}, \textbf{z}_{N,j})
\end{equation}
\begin{equation}
    \text{XAttnDecoder}(\textbf{z}_{N,i}) = {\textbf{z}^{\times}}_{N,i} + \text{MLP}({\textbf{z}^{\times}}_{N,i})
\end{equation}

\subsection{Masking strategies}
\label{subsec:masking strategies}

We propose to study two masking strategies: independent masking and consistent masking. Following MAE \cite{he2022masked}, we apply a 75\% masking ratio and sample our patches uniformly across modalities.

\textbf{Independant masking} - In the MiM for RS literature \cite{cong2022satmae, zhang20233dmae}, independent masking across modalities is widely adopted, as it enables to capture both inter- and intra- modalities correlations. Following this strategy, we randomly sample our masked patches uniformly across modalities.

\textbf{Consistent masking} - We also study a consistent masking strategy, where masked patches are the same across modalities. Our hypothesis is that, given the domain gap between SAR and optical data, capturing inter-modalities correlations is easier than intra-modalities. By guaranteeing that we feed tokens representing the same patches across modalities, we reduce the difficulty for the attention layers to capture cross-modal information.

\section{EXPERIMENTS AND RESULTS}
\label{sec:results}

\begin{table*}[!htb]
    \begin{minipage}{.5\linewidth}
      \centering
            \begin{tabular}{@{}lccc@{}}
            \toprule
                            & S1            & S2            & S1+S2         \\ \hline
            ImageNet        & 70.2          & 85.5          & 83.4          \\
            Dino-MM         & 69.7          & 83.9          & 84.6          \\
            SatViT          & 75.4          & 85.6          & 85.5          \\ \hline
            Fus-MAE XAD     & 75.1          & 86.9          & 87.2          \\
            Fus-MAE XAE     & \textbf{75.9} & \textbf{87.6} & \textbf{88.1} \\ \hline
            \end{tabular}
            \label{tab:results bigearthnet-full}
    \end{minipage}%
    \begin{minipage}{.5\linewidth}
      \centering
            \begin{tabular}{@{}lccc@{}}
            \toprule
                            & S1            & S2            & S1+S2         \\ \hline
            ImageNet        & 55.9          & 58.5          & 60.5          \\
            Dino-MM         & 52.7          & 58.7          & 60.3          \\
            SatViT          & 52.4          & 58.5          & 58.0          \\ \hline
            Fus-MAE XAD     & \textbf{64.8} & \textbf{71.8} & \textbf{72.2}          \\
            Fus-MAE XAE     & 57.5          & 68.0          & 70.0 \\ \hline
            \end{tabular}
            \label{tab:results bigearthnet-1pc}
    \end{minipage} 
    \caption{Mean Average Precision results for the BigEarthNet-MM dataset. Left = finetuning with 100\% of labels. Right = linear evaluation with 1\% labels. S1 = SAR data only. S2 = optical data only. S1+S2= SAR-optical data fusion. XAD = XAttnDecoder, XAE = XAttnEncoder}
    \label{tab:results bigearthnet}
\end{table*}

\begin{table*}[h!]
    \centering
    \begin{tabular}{@{}lccccc@{}}
    \toprule
                    & Top1-Acc      & Top3-Acc      & Precision     & Recall        & F1-score  \\ \hline
    ImageNet        & 58.6          & 90.7          & 75.2          & 58.6          & 60.1      \\
    Dino-MM         & 58.8          & 91.0          & 71.4          & 58.8          & 60.3      \\
    SatViT          & 59.1          & 91.8          & \textbf{75.2} & 59.1          & 59.9      \\ \hline
    Fus-MAE XAD     & 58.8          & 94.0          & 73.9          & 58.7          & 60.5     \\
    Fus-MAE XAE     & \textbf{60.6} & \textbf{94.3} & 71.5          & \textbf{60.6} & \textbf{61.0}             \\ \hline
    \end{tabular}
    \caption{Classification report for the \textbf{linear evaluation} experiment on the SEN12MS dataset. XAD = XAttnDecoder, XAE = XAttnEncoder.}
    \label{tab:results sen12ms}
\end{table*}

\textbf{Benchmark setup} - To study two different fusion strategies, we pretrained two instances of Fus-MAE: Fus-MAE XAE, which performs early-level fusion during encoding, and Fus-MAE XAD, which performes feature-level fusion during decoding. We train our models for 100 epochs on the 354,196 images the the BigEarthNet training split, applying the AdamW optimizer with batch size 200, with a learning rate of $1,5625 \times10^4$. We train our models on 2 NVIDIA RTX 3090Ti, for about 60 hours. We set ImageNet initialization as the baseline, and complete our benchmark with two pretrained Transformer-based models that were specifically designed for SAR-optical data fusion: DINO-MM \cite{wang2022self} and SatVIT \cite{fuller2022satvit}, respectively representing recent studies in contrastive learning and masked image modelling. 

\textbf{Multilabel classification} - To evaluate the effectiveness of our pretraining strategies, we append a linear classifier head on top of our pretrained encoder. We finetune the model over 10 epochs, using a multi-label soft margin loss and the AdamW optimizer, and report the mean average precision (mAP) score over the test split. Table \ref{tab:results bigearthnet} summarizes the outcome of these finetuning experiments, where the classifier is trained on unimodal data (S1 or S2) or multimodal data (S1+S2). On this task, the early fusion approach shows higher performance, showing the potency of our cross-attended patch projection module. It is noteworthy that, across all architectures, the mAP using only SAR data (S1) is significantly lower than when using S2 or S1+S2, suggesting that all models mainly rely on optical data for their predictions in the S1+S2 scenario. Additionally, to assess the quality of the learned representations under label- and resource-scarce conditions, we perform a linear evaluation on 1\% of the training labels. In this scenario, the linear patch projection and the encoder weights are frozen, allowing only the weights of the linear classifier to be learned. We train this classifier for 20 epochs, with a batch size of 128, still utilizing the AdamW optimizer. Results are also reported on Table \ref{tab:results bigearthnet}, revealing an even larger performance increase compared to other SSL architectures, highlighting the quality of the learned representations of our models.

\textbf{Transfer learning} - To study the generalization potential, we perform a linear evaluation on another downstream task: unimodal land-cover classification on the SEN12MS dataset \cite{schmitt2019sen12ms}. We train this classifier for 10 epochs. Given the unbalanced nature of this dataset, we computed the classification metrics using the weighted average method, and applied a label smoothing cross entropy loss. Results of a benchmark with a similar setup as the previous experiment are reported on Table \ref{tab:results sen12ms}. Our models also outperform other techniques across all tracked metrics, although by a thinner margin.

\section{CONCLUSION}
\label{sec:conclusion}

In this paper, we introduce Fus-MAE, a novel SSL framework for SAR-optical data fusion. Based on the MAE architecture, it uses cross-attention between two data streams at different stages to perform early and feature-level data fusion. Our model outperforms recent contrastive learning and MIM-based works on various downstream tasks, demonstrating the effectiveness of using cross-attention to describe cross-modal interactions between modalities with a large domain gap.
Further research can be conducted to adapt our cross-attention layers to more than 2 modalities and to balance the prediction reliance of our model more evenly between the modalities.  

\bibliographystyle{IEEEbib}
\bibliography{strings,refs}

\end{document}